\documentclass[conference]{IEEEtran}

\usepackage{times}

\usepackage{graphicx}
\usepackage{epsfig}
\usepackage{subfigure}
\usepackage[cmex10]{amsmath}
\usepackage{epstopdf}
\usepackage{multirow}
\usepackage{array}
\usepackage{cite}
     \usepackage{balance}
     \usepackage{algorithmic}
 \usepackage{algorithm}

\begin{document}
\title{Facial Emotion Recognition using Min-Max Similarity Classifier}

\author{\IEEEauthorblockN{Olga Krestinskaya and Alex Pappachen James}
\IEEEauthorblockA{
School of Engineering, Nazarbayev University, Astana\\
www.biomicrosystems.info/alex\\
Email: apj@ieee.org}}

\maketitle

\begin{abstract}

Recognition of human emotions from the imaging templates is useful in a wide variety of human-computer interaction and intelligent systems applications. However, the automatic recognition of facial expressions using image template matching techniques suffer from the natural variability with facial features and recording conditions. In spite of the progress achieved in facial emotion recognition in recent years, the effective and computationally simple feature selection and classification technique for emotion recognition is still an open problem. In this paper, we propose an efficient and straightforward facial emotion recognition algorithm to reduce the problem of inter-class pixel mismatch during classification. 
The proposed method includes the application of pixel normalization to remove intensity offsets followed-up with a Min-Max metric in a nearest neighbor classifier that is capable of suppressing feature outliers. The results indicate an improvement of recognition performance from 92.85\% to 98.57\% for the proposed Min-Max classification method when tested on JAFFE database. The proposed emotion recognition technique outperforms the existing template matching methods.

\end{abstract}








\section{Introduction}

In recent years, the human-computer interaction challenge has led to the demand to introduce efficient facial and speech recognition systems \cite{els1,els2, els3, els5, b1}. Facial emotion recognition is the identification of a human emotion based on the facial expression and mimics \cite{DWT}. The facial emotion recognition has a wide range of appliction prospects in different areas, such as medicine \cite{med}, robotics \cite{3,appl1}, computer vision, surveillance systems \cite{els1}, machine learning \cite{appl4}, artificial intelligence, communication \cite{appl2,mobile}, psychological studies \cite{els5}, smart vehicles \cite{appl1}, security and embedded systems \cite{appl3}. 

There are two main approaches for facial expression recognition: geometry-based and appearance-based methods \cite{els2}. The geometry-based methods extract main feature points and their shapes from the face image and calculate the distances between them. While, appearance-based methods focus on the face texture using different classification and template matching methods \cite{sth2,sth3}. In this paper, we focus on facial emotion recognition based on template matching techniques that remains a challenging task \cite{new1,new2,int1}. Since the orientation of pixel features are sensitive to the changes in illumination, pose, scale and other natural imaging variabilities, the matching errors tend to be high \cite{els5, els4, ref1}. Pixel matching methods are known to be useful when the images has missing features because imaging matrices become sparse and feature computation process is not trivial. As facial expressions cause a mismatch of intra-class features due to their orientation variability, it is difficult to map them between the imaging templates. 

Facial emotion recognition accuracy depends on the robustness of a feature extraction method to intra-class variations and classifier performance under noisy conditions and with various types of occlusions \cite{appl4}. Even thought a variety of approaches for the automated recognition of human expressions from the face images using template matching methods have been investigated and proposed over the last few years \cite{sth2}, the emotion recognition method with robust feature extraction and effective classification techniques accompanied by low computational complexity is still an open research problem \cite{sth}. Therefore, in this paper, we address the issues
of matching templates through pixel normalization followed by the removal of inter-image feature outliers using a Min-Max similarity metric. We apply Gaussian normalization method with local mean and standard deviation to normalize the pixels and extract relevant face features followed by Min-Max classification method to determine an emotion class. The simulation is performed in Matlab for the Japanese Female Facial Expression (JAFFE) database \cite{8} and the emotion recognition accuracy is calculated using leave-one-out cross-validation method.

The main contributions of this work are the following:
\begin{itemize}
\item We develop a simplified approach for facial emotion recognition with template matching method using Gaussian normalization, mean and standard deviation based feature extraction and Min-Max classification approach.
\item We present simple and effective facial emotion recognition algorithm having low computational complexity and ability to suppress the outliers and remove intensity offsets.
\item We conduct the experiments and simulations on JAFFE database to demonstrate the efficiency of the proposed approach and highlight its advantages, comparing to the other existing methods. 

\end{itemize}

The paper is organized as follows. Section \ref{sec2} presents the overview of the existing methods for facial emotion recognition, their drawbacks and reasons to propose a new method. In Section \ref{sec3}, we show normalization, feature extraction and classification parts of the proposed method, present the algorithm and describe the experiments. Section \ref{sec4} contains the simulation results and comparison of the obtained results with the existing methods. In Section \ref{sec5}, we discuss the benefits and drawbacks of the proposed method, in comparison to the traditional methods. Section \ref{sec6} concludes the paper.

\section{Background and related works}
\label{sec2}

To address the problem of facial emotion recognition, several template matching methods have been proposed in the last decades \cite{els1, 3,4,5,comp1}. In most of the cases, the process of emotion recognition from human face images is divided into two main stages: feature extraction and classification \cite{els1,3}. The main aim of feature extraction methods is to minimize intra-class variations and maximize inter-class variations. The most important facial elements for human emotion recognition are eyes, eyebrows, nose, mouth and skin texture. Therefore, a vast majority of feature extraction methods focus on these features \cite{els2,p1}. The selection of irrelevant face image features or insufficient number of them would lead to low emotion recognition accuracy, even applying effective classification methods \cite{sth}. The main purpose of the classification part is to differentiate the elements of different emotion classes to enhance emotion recognition accuracy. 

The commonly used feature extraction methods include two-dimensional Linear Discriminant Analysis (2D-LDA) \cite{3,comp1}, two-dimensional Principle Component Analysis (2D-PCA) \cite{PCA}, Discrete Wavelet Transform
(DWT) \cite{SVM2,3,DWT}, Gabor based methods \cite{els6,els7} and wavelets-based techniques \cite{4,Ahmad2}. In 2D-LDA method, the two-dimensional image matrix is exploited to form scatter matrices between the classes and within the class \cite{3}. 2D-LDA method can be applied for facial features extraction alone or accompanied with the other feature extraction method, as DWT \cite{3,comp1}. In 2D-PCA feature extraction method, the covariance matrix representation of the image is derived directly from the original image \cite{3,PCA}. The size of the derived principle component matrix is smaller than the original image size that allows to decrease the amount of processing data, and consequently, reduce the required computational memory \cite{PCA2}. However, 2D-LDA and 2D-PCA methods applied in template matching techniques require an additional processing of the image, dimensionality reduction techniques or application of another feature extraction method to achieve higher recognition accuracy, which leads to the increase in processing time.

The other feature extraction method is DWT. This method is based on the low-pass and high-pass filtering, therefore, it is appropriate for the images with different resolution levels \cite{3}. In the emotion recognition task, DWT is applied for the extraction of useful features from the face images and can be replaced with its Orthogonal Wavelet Transform (OWT) and Biorthogonal Wavelet Transform (BWT) having the advantages of orthogonality \cite{DWT}. Another method for facial emotion recognition is Gauss-Laguerre wavelet geometry based technique. This method represents the processed image in polar coordinates with the center at a particular pivot point. The degree of freedom is one of the advantages that Gauss-Laguerre approach provides, which in turn allows to extract of features of the desirable frequency from the images \cite{4,Ahmad2}. However, DWT and Gauss-Laguerre approaches are complex and require time and memory consuming calculations. 

The classification of the extracted features can be implemented using Support Vector Machine (SVM) algorithm \cite{3,SVM2}, K-Nearest Neighbor (KNN) method \cite{4,KNN}, Random Forest classification method \cite{med,forest1} and Gaussian process \cite{5}. The SVM principle is based on non-linear mapping and identification of a hyperplane for the separation of data classes. SVM classifier is used with the application of different kernel functions, such as linear,
quadratic, polynomial and radial basis functions, to optimize the SVM performance \cite{3,SVM2}. KNN approach is based on the numerical comparison of a testing data sample with training data samples of each class followed by the determination of the similarity scores. The data class is defined by K most similar data samples based on the minimum difference between train and test data \cite{4,KNN}. KNN and SVM classifiers are simple and widely used for emotion recognition, however these classifiers do not suppress the outliers that leads to lower recognition accuracy.

The Random Forest classification method is based on the decision making tree approach with randomized parameters \cite{med,forest1}. To construct the decision tree, Random Forest algorithm takes a random set of options and selects the most suitable from them \cite{forest2}. Random Forest classifier is robust and has a high recognition rate for the images of large resolution \cite{forest3}. The drawback of Random Forest classifier is its computational complexity. The other classification method is the Gaussian process approach. Gaussian process is based on the predicted probabilities and can be used for facial emotion recognition without application of feature selection algorithms. The Gaussian process allows a simplified computational approach, however has a smaller emotion recognition rate, comparing to the other methods \cite{5}. 

Even thought the a number of methods for feature extraction and classification have been proposed, there is a lack of template matching methods that allow to achieve high recognition accuracy with minimum computational cost. Therefore, the method that we propose has a potential to reduce the computational complexity of facial emotion recognition operation and increase the recognition accuracy due to the simplicity of the algorithm, effective feature extraction and ability to suppress outliers. The proposed algorithm can be implemented using small computational capacity devices keeping facial emotion recognition operation fast and accurate.

\section{Methodology}
\label{sec3}
\begin{figure}[!h]
\centering{\includegraphics[width=70mm]{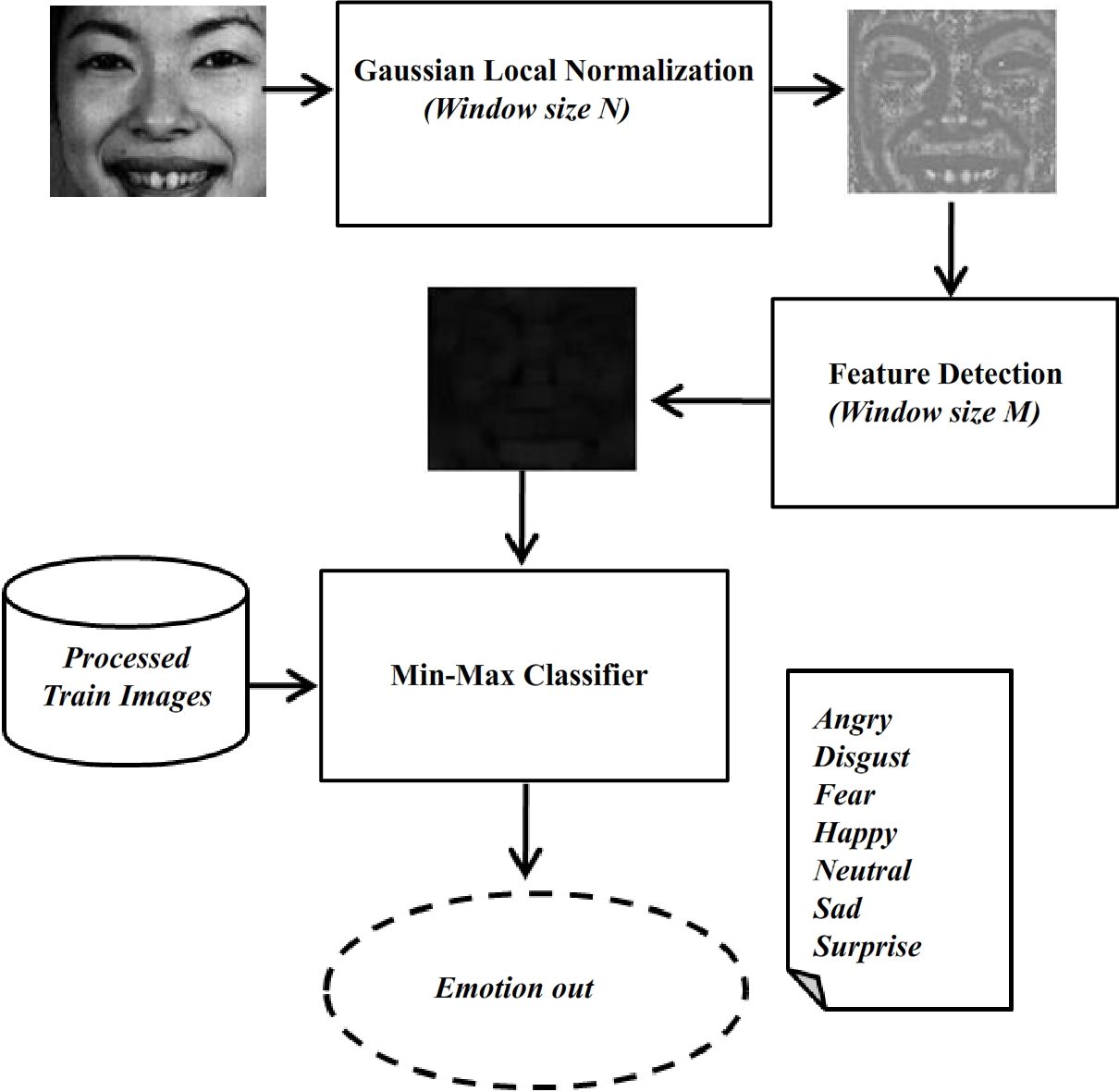}}
\caption{Outline of the proposed emotion recognition system
}
\label{1}
\end{figure}

The main idea of the proposed method is to extract the spatial change of standardized pixels in a face image and detect the emotion class of the face using a Min-Max similarity Nearest Neighbor classier. The images from the JAFFE database \cite{8} are used for the experiments.  This database contains 213 images of 10 female faces comprising 6 basic facial expressions and neutral faces. The original images from the database have the size of $256\times256$ pixels and in our experiments they are cropped to a size of $101\times114$ pixels retaining only the relevant information of the face area. A block diagram of the proposed method is shown in Fig. \ref{1}. 

\subsection{Pre-processing}

Illumination variability introduces the inter-class feature mismatch resulting in the inaccuracies in the detection of emotion discriminating features from the face images. Therefore, image normalization is essential to reduce the inter-class feature mismatch that can be viewed as intensity offsets. Since the intensity offsets are uniform within a local region, we perform Gaussian normalization using local mean and standard deviation. The input image is represented as $x(i, j)$, and $y(i,j)$ is the normalized output image, where $i$ and $j$ are the row and column number of the processed image.  The normalized output image is calculated by Eq. \ref{Eq1} \cite{james2010inter}, where $\mu$ is a local mean and $\sigma$ is a local standard deviation computed over a window of $N\times N$ size.

\begin{align}\label{Eq1}
\begin{split}
y(i,j)=\frac{x(i,j)-\mu(i,j)}{6\sigma(i,j)}
\end{split}
\end{align}

The parameters $\mu$ and $\sigma$ are calculated using Eq. \ref{Eq2} and \ref{Eq3} with $a=(N-1)/2$. 

\begin{align}\label{Eq2}
\begin{split}
\mu(i,j)=\frac{1}{N^2}\sum_{k=-a}^{a} \sum_{h=-a}^{a}x(k+i,h+j)
\end{split}
\end{align}

\begin{align}\label{Eq3}
\begin{split}
\sigma(i,j)=\sqrt{\frac{1}{N^2}\sum_{k=-a}^{a} \sum_{h=-a}^{a}{[x(k+i,h+j)-\mu(i,j)]^2}}
\end{split}
\end{align}

\begin{figure}[h]
\centering
 \begin{subfigure}[]{}
              \includegraphics[width=43mm]{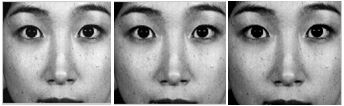}
        \end{subfigure}%
         \begin{subfigure}[]{}
         \includegraphics[width=43mm]{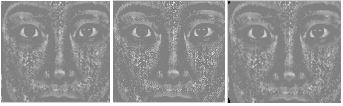}
        \end{subfigure}%
         \begin{subfigure}[]{}
         \includegraphics[width=43mm]{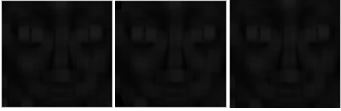}
        \end{subfigure}%
        \caption{(a) Sample image from JAFFE database with different lighting conditions obtained by adding and subtracting a value of 10 from the original image.(b) Normalized images of above sample images obtained by performing Gaussian normalization using local mean and local standard deviation taken over a window of size N=11. (c) Feature detected images from the normalized image by performing local standard deviation using a window of size M=11.}
          \label{fig2a} 
\end{figure}

An example image from the JAFFE database with three different lighting conditions is shown in Fig. \ref{fig2a} (a). As the JAFFE database does not contain the face images with different illumination conditions, the illumination change was created by adding and subtracting the value of 10 from the original image. Fig. \ref{fig2a} (b) illustrates the respective images after Gaussian local normalization. Irrespective of the illumination conditions, the three locally normalized images appear similarly with the minimum pixel intensity variation.  

\subsection{Feature detection}

The feature parts useful for the facial emotion recognition are eyes, eyebrows, cheeks and mouth regions. In this experiment, we perform the feature detection by calculating local standard deviation of normalized image using a window of M$\times$M size. Eq. \ref{Eq4} is applied for the feature detection with $b=(M-1)/2$.

\begin{align}\label{Eq4}
\begin{split}
w(i,j)=\sqrt{\frac{1}{M^2}\sum_{k=-b}^{b} \sum_{h=-b}^{b}{[y(k+i,h+j)-\mu'(i,j)]^2}}
\end{split}
\end{align}

In Eq. \ref{Eq4} the parameter $\mu'$ refers to the mean of the normalized image $y(i,j)$ and can be calculated by Eq. \ref{Eq5}.

\begin{align}\label{Eq5}
\begin{split}
\mu'(i,j)=\frac{1}{M^2}\sum_{k=-b}^{b} \sum_{h=-b}^{b}y(k+i,h+j)
\end{split}
\end{align}

Fig. \ref{fig2a} (c) shows the results of feature detection corresponding to the normalized images.

%


\subsection{Emotion Classification}

 For the recognition stage, we propose a Min-Max similarity metric in a Nearest Neighbor classifier framework. This method is based on the principle that the ratio of the minimum difference to the maximum difference of two pixels will produce a unity output for equal pixels and an output less than unity for unequal pixels. The proposed method is described in Algorithm \ref{alg1}. The algorithm parameter $trainlen$ refers to the number of train images, $N$ corresponds to the normalization window size, and $M$ indicates the feature detection window size. 
Each cropped image is of $m\times n$ pixel dimension. The parameter \textit{train} is a feature array of $trainlen\times$($m\times n$) size, where each row corresponds to the processed train images. After normalization and feature detection, test images are stored in to a vector \textit{test} of $ 1\times$($m\times n$) size. A single test image is compared pixel-wise with processed train images of all the classes in the feature array using the proposed Min-Max classifier:

\begin{align}\label{Eq6}
\begin{split}
s(i,j)=[\frac{min[train(i,j),test(1,j)]}{max[train(i,j),test(1,j)]}]^\alpha,
\end{split}
\end{align}

where a parameter $\alpha$ controls the power of exponential to suppress the outlier similarities. Outliers are the observations that come inconsistent with the remaining observations and are common in real-time image processing. The presence of outliers may cause misclassification of an emotion, since sample maximum and sample minimum are maximally sensitive to them. In order to remove the effects of the outliers, $\alpha=3$ is selected to introduce the maximum difference to the inter-class images and minimum difference to the intra-class images. 

After Min-Max classification, a column vector $z$ of $trainlen \times 1$ size containing the weights obtained after comparing the test image with each of the \textit{trainlen} number of train images is calculated using Eq. \ref{Eq7}.

\begin{align}\label{Eq7}
\begin{split}
z(i)=\sum_{j=1}^{m\times n}s(i,j)
\end{split}
\end{align}

The classification output \textit{out} shown in Eq. \ref{Eq8} is the maximum value of $z$ corresponding to the train image that shows the maximum match.  The recognized emotion class is the class of the matched train image.

\begin{align}\label{Eq8}
\begin{split}
out=max(z(i))
\end{split}
\end{align}

\begin{algorithm}[!h]
\caption{Emotion Recognition using Min-Max classifier }
\label{alg1}
\begin{algorithmic}[1]
\REQUIRE {Test image Y,Train images $X_t$,trainlen,window size N and M}

\STATE {Crop the images to a dimension of $m\times n$}

\FOR{$t=1$ to $trainlen$}

\STATE $C(i,j)=\frac{X_t(i,j)-\mu(i,j)}{6\sigma(i,j)}$

\STATE $W(i,j)=\sqrt{\frac{1}{M^2}\sum_{k=-b}^{b} \sum_{h=-b}^{b}{[C(k+i,h+j)-\mu(i,j)]^2}}$

\STATE {Store the value of W to an array $train$ of dimension $trainlen\times m\times n$}

\ENDFOR

\FOR{$t=1$ to $trainlen$}

\STATE $V(i,j)=\frac{Y(i,j)-\mu(i,j)}{6\sigma(i,j)}$

\STATE $test(i,j)=\sqrt{\frac{1}{M^2}\sum_{k=-b}^{b} \sum_{h=-b}^{b}{[V(k+i,h+j)-\mu(i,j)]^2}}$ 

\STATE $s(t,j)=[\frac{min[train(t,j),test(1,j)]}{max[train(t,j),test(1,j)]}]^3$

\STATE $z(t)=\sum_{j=1}^{m\times n}s(t,j)$

\STATE $ out=max(z(t))$
\ENDFOR

\end{algorithmic}
\end{algorithm}

\section{Results and comparison }
\label{sec4}
\begin{figure}[!h]

\centering{\includegraphics[width=80mm]{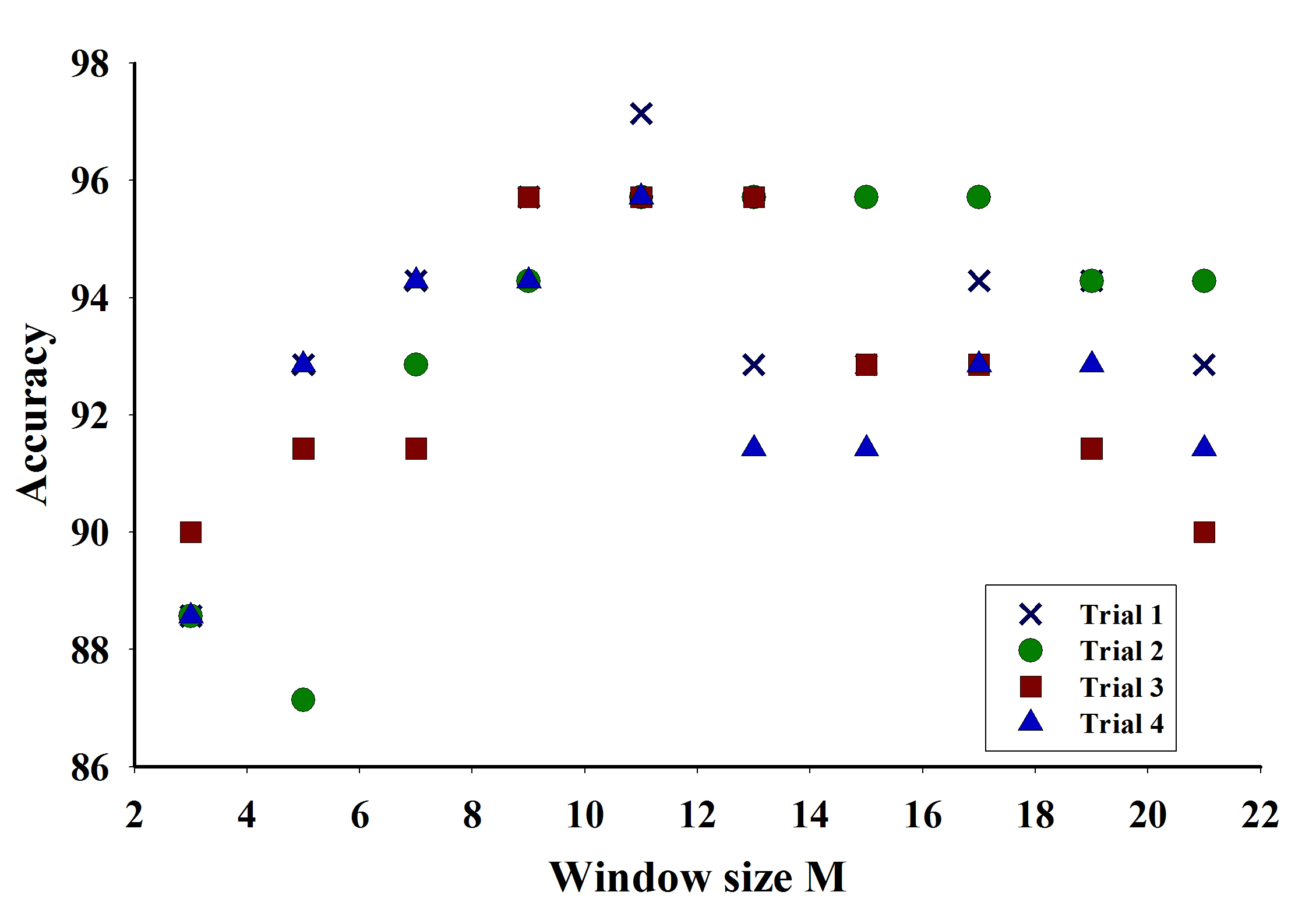}}
\caption{The accuracy rates obtained for four trials of leave-one-out cross-validation method for different feature detection window size M ranging from 3 to 21.}
\label{ff3}
\end{figure}

To benchmark the performance of the proposed algorithm, leave-one-out cross-validation method has been used. In this method, one image of each expression of each person is applied for testing and the remaining images are used for training \cite{4}. The cross-validation is repeated 30 times to obtain a statistically stable performance of the recognition system and to ensure that all the images in JAFFE database are used for testing at least once. The overall emotion recognition accuracy of the system is obtained by averaging the results of the cross-validation trials. Fig. \ref{ff3} shows the different accuracy rates obtained for each trial by varying feature detection window size $M$ from 3 to 21 and keeping normalization window size at $N=11$. It is shown that the maximum emotion recognition accuracy this normalization window size can be achieved with the detection window size of 11.

\begin{figure}[!h]
\centering{\includegraphics[width=90mm]{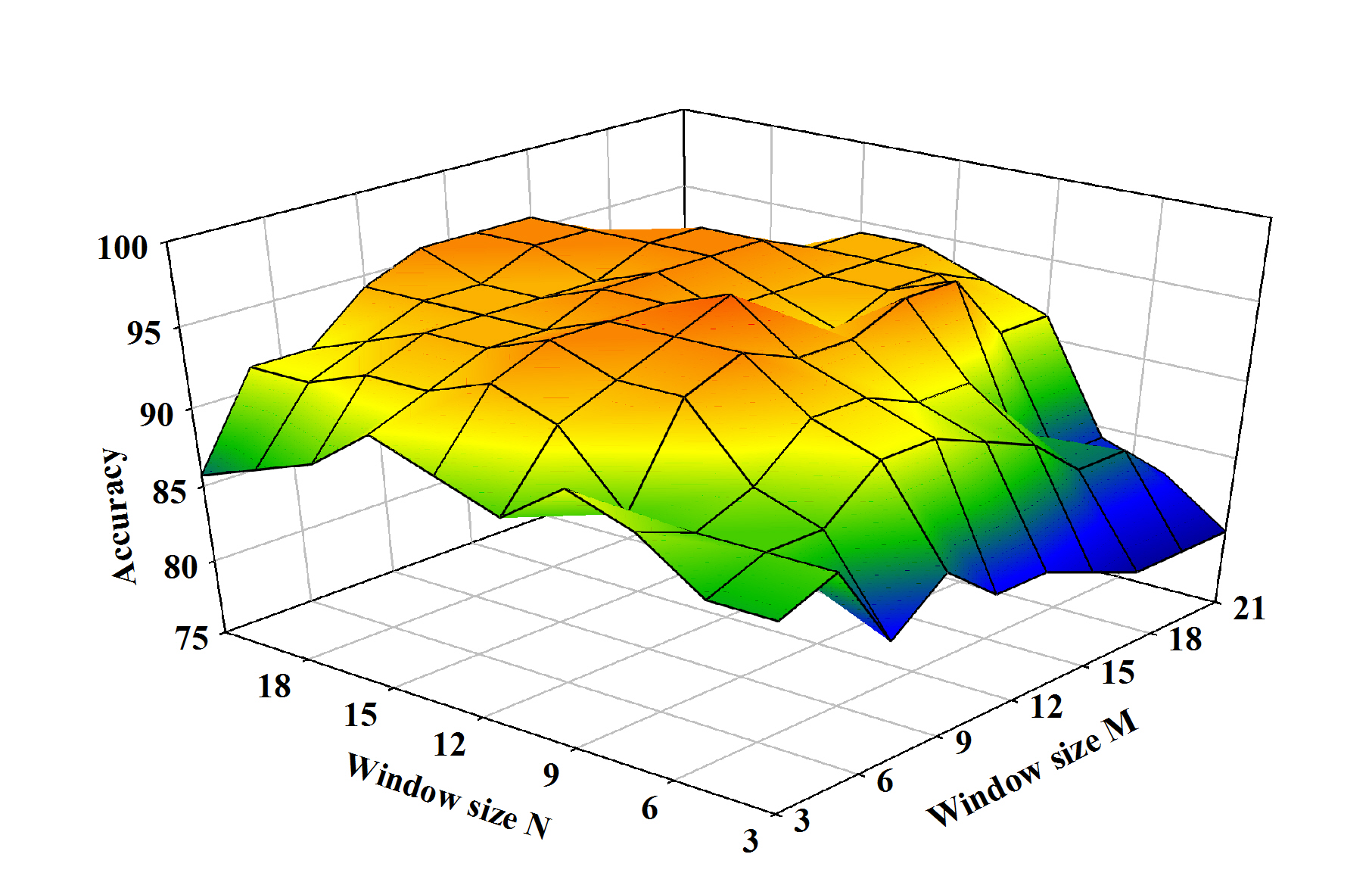}}
\caption{Graph shows the accuracy rates obtained for different normalization and feature detection window sizes ranging from M=N=3 to 21 for one trial of leave-one-out cross-validation method.}
\label{f4}
\end{figure}

Fig. \ref{f4} shows the recognition accuracy rates obtained for different normalization and feature detection window sizes ranging from 3 to 21 for a single trial. 

To evaluate the performance of the proposed Min-Max classifier, it has been compared with the other classifiers, such as Nearest Neighbor\cite{6} and Random Forest\cite{7}, after normalization and feature detection. The obtained accuracy values are shown in Table \ref{tb1}.

\begin{table}[!h]
\centering
 \caption{Testing of feature detected images on other classifiers}
 \begin{tabular}{l|c}\hline
 $Classifier$ & $Accuracy (\%)$ \\\hline
 Nearest Neighbor & 92.85 \\
 Random Forest & 88.57 \\
 \textbf{Proposed Min-Max classifier} & \textbf{98.57}\\\hline
 \end{tabular}
 \label{tb1}
 \end{table}

 The proposed method achieves a maximum accuracy of 98.57\% for a window size of $M=N=11$ which outperforms the other existing methods in the literature for leave-one-out cross-validation method. Table \ref{tb2} shows the performance comparison of the proposed method with the other existing systems on the same database using leave-one-out cross-validation method. Applying the proposed method, we achieved emotion recognition accuracy of 98.57\% proving the significance of emotion detection method with Min-Max similarity classifier.

\begin{table}[!h]
 \caption{Comparison of proposed emotion recognition system with other existing system based on leave-one-out cross-validation method}
 
{\begin{tabular}{|l|p{2.5cm}|c|}\hline
$Existing$ $systems$ & $Method$ $used$ & $Accuracy (\%)$\\\hline
Cheng et. al \cite{5} & Gaussian Process & 93.43 \\\hline
Hamester et. al \cite{cnn} & Convolutional Neural Network & 95.80 \\\hline
Frank et. al \cite{3} & DWT + 2D-LDA +SVM & 95.71 \\\hline
Poursaberi et. al \cite{4} & Gauss Laguerre wavelet+ KNN & 96.71\\\hline

Hegde et. al \cite{els6}& Gabor and geometry based features & 97.14
\\\hline

\textbf{Proposed method} & \textbf{Min-Max classifier} & \textbf{98.57}\\\hline
\end{tabular}}{}
\label{tb2}

\end{table}
In addition, comparing to the proposed emotion recognition system, the other existing methods require specialized feature extraction and dimensionality reduction techniques before classification stage. The main advantages of the proposed emotion recognition system are its simplicity and straightforwardness.
\section{Discussion}
\label{sec5}

The main advantages of the proposed facial motion recognition approach are high recognition accuracy and low computational complexity. To achieve high recognition accuracy, the effective feature selection is required. In the existing methods, the complex algorithms for feature selection are applied without normalizing the image. The normalization stage is important and has a considerable effect on the accuracy of the feature selection process. In the proposed algorithm, we apply simple pre-processing methods to normalize the images and eliminate intensity offsets that effects the accuracy of the feature selection process and leads to the increase of emotion recognition accuracy, in comparison to the existing methods. The effect of the proposed Min-Max classification method on recognition accuracy is also important. Table \ref{tb1} shows the application of the other classification method for the same approach where proposed Min-Max classifier illustrates the performance improvement. In comparison to the existing method, the proposed Min-Max classifier has an ability to suppress the outliers that significantly impacts overall performance of this approach.

The simplicity and straightforwardness of the proposed approach are also important due to the resultant low computational complexity. Most of the existing methods use complicated feature extraction and classification approaches that double the complexity of the facial recognition process and require the device with large computational capacity to process the images. We address this problem applying direct local mean and standard deviation based feature detection methods and simple Min-Max classification method. In comparison to the existing feature detection methods, such as PCA \cite{PCA} and LDA \cite{3}, the proposed method is straightforward and does not require dimensionality reduction. Moreover, simple Min-Max classification method also reduce the computational time, comparing to the traditional classification approaches, such as SVM \cite{SVM2}, KNN \cite{KNN}, Gaussian process \cite{5} and Neural Network \cite{nn}. Therefore, the algorithm can be run on the device with low computational capacity.


\section{Conclusion}
\label{sec6}

In this paper, we have represented the approach to improve the performance of emotion recognition task using template matching method. We have demonstrated that the pixel normalization and feature extraction based on local mean and standard deviation followed up by the Mix-Max similarity classification can result in the improvement of overall classification rates. We achieved emotion recognition accuracy of 98.57\% that exceeds the performance of the existing methods for the JAFFE database for leave-one-out cross-validation method.  The capability of the algorithm to suppress feature outliers and remove intensity offsets results in the increase of emotion recognition accuracy. Moreover, the proposed method is simple and direct, in comparison to the other existing methods requiring the application of dimensionality reduction techniques and complex classification methods for computation and analysis. Low computational complexity is a noticeable benefit of the proposed algorithm that implies the reduction of computational time and required memory space. This method can be extended to the other template matching problems, such as face recognition and biometric matching. The drawback of the proposed method, as in any other template matching method, is the metric learning requiring to create the templates for each class that, in turn, consumes additional memory space to store the templates.  

\bibliographystyle{IEEEtran}
\bibliography{mybibfile.bib}

\end{document}